\documentclass[letterpaper, 10 pt, conference]{ieeeconf}  

\IEEEoverridecommandlockouts                              

\usepackage[pdftex]{graphicx}
\usepackage{times}
\usepackage{epsfig}
\usepackage{amsfonts}
\usepackage{amsmath}
\usepackage{amssymb}
\usepackage{adjustbox}
\usepackage{threeparttable}
\usepackage{xcolor}
\usepackage{authblk}
\usepackage{multirow}
\usepackage{caption}
\usepackage{float}
\usepackage{bm}
\usepackage{bbm}
\usepackage{booktabs}
\usepackage{gensymb}
\usepackage{mathtools}
\usepackage{textcomp}
\usepackage{tabularx}
\usepackage{makecell}
\usepackage{array}
\usepackage{siunitx}
\usepackage{xspace}
\usepackage{cite}
\usepackage{url}
\usepackage{hyperref}

\makeatletter
\DeclareRobustCommand\onedot{\futurelet\@let@token\@onedot}
\def\@onedot{\ifx\@let@token.\else.\null\fi\xspace}

\def\eg{\emph{e.g}\onedot}

\def\etc{\emph{etc}\onedot} 
 
\def\etal{\emph{et al}\onedot}
\makeatother

\newcommand{\norm}[1]{\left\lVert#1\right\rVert_2}

\begin{document}

\title{SAGE: SLAM with Appearance and Geometry Prior for Endoscopy}

\author{
Xingtong~Liu, Zhaoshuo~Li, Masaru~Ishii, Gregory~D.~Hager,~\IEEEmembership{Fellow,~IEEE,} \\
Russell~H.~Taylor,~\IEEEmembership{Life Fellow,~IEEE,} and Mathias~Unberath
\thanks{Xingtong~Liu was, Zhaoshuo~Li, Gregory~D.~Hager, Russell~H.~Taylor, and Mathias~Unberath are with the Computer Science Department, Johns Hopkins University (JHU), Baltimore, MD 21287 USA.}
\thanks{Masaru Ishii is with Johns Hopkins Medical Institutions, Baltimore, MD 21224 USA.}
\thanks{Under a license agreement between Galen Robotics, Inc and JHU, Dr. Taylor and JHU are entitled to royalty distributions on technology related to this publication.
Dr. Taylor also is a paid consultant to and owns equity in Galen Robotics, Inc.
This arrangement has been reviewed and approved by JHU in accordance with its conflict-of-interest policies.}
\thanks{This work was supported in part by a fellowship from Intuitive Surgical.}
}

\maketitle
\thispagestyle{empty}
\pagestyle{empty}

\begin{abstract}
In endoscopy, many applications (\eg,~surgical navigation) would benefit from a real-time method that can simultaneously track the endoscope and reconstruct the dense 3D geometry of the observed anatomy from a monocular endoscopic video.
To this end, we develop a Simultaneous Localization and Mapping system by combining the learning-based appearance and optimizable geometry priors and factor graph optimization.
The appearance and geometry priors are explicitly learned in an end-to-end differentiable training pipeline to master the task of pair-wise image alignment, one of the core components of the SLAM system.
In our experiments, the proposed SLAM system is shown to robustly handle the challenges of texture scarceness and illumination variation that are commonly seen in endoscopy.
The system generalizes well to unseen endoscopes and subjects and performs favorably compared with a state-of-the-art feature-based SLAM system.
The code repository is available at \href{https://github.com/lppllppl920/SAGE-SLAM.git}{https://github.com/lppllppl920/SAGE-SLAM.git}.
\end{abstract}

\IEEEpeerreviewmaketitle

\section{Introduction}
Endoscopy is a technique allowing inspection, manipulation, and treatment of internal organs using devices from a distance of the target organs without a large incision.
Nowadays, the quality of an endoscopic procedure is directly related to the attitude and level of skills of the person who drives the endoscope~\cite{groen2017history}.
When inspection or surgeries are performed, there is a risk of iatrogenic perforations~\cite{Fockens2016perforation}.
In cases where critical structures below the surface get damaged, the consequence can be detrimental.
One of these is endoscopic endonasal surgery (ESS), which requires a thorough knowledge of anatomy, in particular, the relationship of the nose and sinuses to adjacent vulnerable structures such as the orbit or base of the skull.
However, malformations, previous operations, and massive polyposis may interfere greatly with the intra-operative orientation of surgeons and this leads to major risks, such as loss of vision, diplopia, injury to the carotid artery, \etc, for patients~\cite{Eliashar2003navigation}.
Having a surgical navigation system that tracks the endoscope and shows the spatial relationship between the scope and the surrounding anatomy can greatly reduce the risk.

Many marker-based navigation systems have been developed and commercialized to provide such information.
Nevertheless, visual-based navigation systems have been preferred compared to marker-based ones because the former do not interrupt the clinical workflow and is robust to the relative movement between the observed anatomy and the patient.
One critical component of such a system is to track the endoscope and estimate the geometry of the observed anatomy from a video stream.
The surface geometry from a video can be aligned with a pre-operative model, \eg,~one from Computed Tomography (CT), with a registration method.
The spatial relationship between the endoscope and the surrounding structures will then be known.
A typical choice for endoscope tracking is a Simultaneous Localization and Mapping (SLAM) system.
Many systems~\cite{grasa2013visual,mahmoud2017slam,wang2019visual} only provide sparse geometry, which mainly serves as a map to track the endoscope but is not sufficient to register against the pre-operative model and also not useful for other clinical applications (\eg,~anatomical shape analysis).
The accuracy and robustness of such systems are also limited in endoscopy because of the scarce textures that lead to less repeatable keypoint detections across frames.
For previous works that estimate dense geometry~\cite{Ma2021rnnslam}, the accuracy of the estimated surface models is not evaluated and the generalizability of such a system on unseen subjects is unknown.

In this work, to robustly track the endoscope and obtain accurate surface geometry of the observed anatomy with a monocular endoscope, we develop a SLAM system that combines the expressivity of deep learning and the rigorousness of non-linear optimization.
Specifically, we exploit learning-based appearance and optimizable geometry priors and factor graph optimization.
Based on our evaluation, the proposed SLAM system generalizes well to unseen endoscopes and subjects and performs favorably compared with a state-of-the-art feature-based SLAM system~\cite{ORBSLAM32020}.
The contributions of this work are as follows: 1) A SLAM system with learning-based appearance and geometry priors for monocular endoscopy. 2) An end-to-end training pipeline to explicitly learn the appearance and geometry priors that are suitable for handling the task of pair-wise image alignment.

\section{Related Work}
\subsection{Representation Learning for Visual Tracking and Mapping}
In recent years, researchers have worked on exploiting prior information learned from data to improve the performance of SLAM and Visual Odometry (VO).
Different forms of depth priors have been used, such as fixed depth estimate~\cite{Tateno2017cnnslam,ma2019real,Ma2021rnnslam}, self-improving depth estimate~\cite{tiwari2020pseudo}, depth estimate with uncertainty~\cite{yang2020d3vo}, and depth estimate with optimizable code~\cite{Bloesch2018codeslam,tang2018ba,Czarnowski2020deepfactors}.
Appearance priors have been studied to replace the role of color images in vision-based methods, which enlarges the convergence basin of optimization and enables scenarios with no photometric constancy.
BA-Net~\cite{tang2018ba} proposed representation learning with differentiable BA-related loss. 
DeepSFM~\cite{wei2020deepsfm} extracted implicit representation with joint depth and pose estimation.
In this work, we integrate both appearance and optimizable depth priors into the SLAM system.
There are also works exploiting other forms of priors for the VO and SLAM systems. For example, Yang~\etal~\cite{yang2020d3vo} exploit a pose prior to enable better convergence and mitigate the scale-drift issue; 
Zhan~\etal~\cite{Zhan2020sz} estimate dense optical flow to gain more robustness towards camera tracking.

\subsection{Simultaneous Localization and Mapping in Endoscopy}
Many SLAM systems have been developed for the general scene~\cite{Engel2014lsdslam,mur2015orb,Cadena2016Past,Tateno2017cnnslam,Artal2017orbslamv2,Li2018Evolution,Bloesch2018codeslam,lamarca2019defslam,Laidlow2019DeepFusion,tiwari2020pseudo,Greene2020metrically,Czarnowski2020deepfactors,ORBSLAM32020}.
In endoscopy, additional challenges exist compared with other scenarios such as driving scenes, which are illumination changes, scarce textures, deformation, \etc.
Feature-based SLAM~\cite{grasa2013visual,mahmoud2017slam,Mahmoud2019LiveTA} has been developed for its robustness to illumination changes.
To deal with the scarce texture that causes inaccurate estimates, works have been proposed using either hardware~\cite{Qiu2018EndoscopeOral} or algorithmic~\cite{Turan2018UnsupervisedOA,liu2020extremely,Ma2021rnnslam} solutions.
Deformation happens in endoscopy, especially in certain cases such as laparoscopy and when surgical operations are applied, and there are works developed to confront this challenge~\cite{Turan2017nrslamendo,song2018mis,Song2019AnOT,lamarca2019defslam}.
In this work, we exploit deep priors and dense geometry to improve the robustness of the system to illumination changes and scarce texture.

\section{Representation Learning}

\subsection{Network Architecture}\label{sec:dense_slam_network_arch}
Two separate networks are used to learn geometry and appearance representations, respectively.
In terms of geometry, a depth network produces an average depth estimate, which is correct up to a global scale, and depth bases.
The average depth estimate captures the expectation of the depth estimate based on the input color image.
However, the task of depth estimation from a single image is ill-posed and therefore errors are expected.
The depth bases consist of a set of depth variations that could be used to explain the variation of geometry given the appearance of the input.
Such bases provide a way to further refine the depth estimate, with additional information, during the SLAM optimization.
The network is close to UNet~\cite{Ronneberger2015unet} with partial convolution~\cite{liu2018image}, where an endoscope mask is used so that blank regions do not contribute to the final output.
There are two output branches, where one, with absolute as output activation, predicts the average depth estimate, and the other produces depth bases with hyperbolic tangent as output activation.
Please refer to the code repository for the architecture of the depth network used for depth training.

In terms of appearance, a feature network produces two sets of representations.
One set, named descriptor map, is used as descriptors in pair-wise feature matching that are involved in the Reprojection Factor and Sparse Matched Geometry Factor, described in Sec.~\ref{sec:dense_slam_factor_design}.
A similar training approach as~\cite{liu2020extremely} is used.
The other set, named feature map, is used for the computation of the Feature-metric Factor as a drop-in replacement of the color image.
This is because, in the image, the illumination of the same location of the scene changes as the viewpoint varies, which is caused by the lighting source moving with the camera.
On the other hand, feature maps can be robust to illumination and viewpoint changes, if the feature network is trained correspondingly.
In this work, we use the task of pair-wise image alignment with differentiable non-linear optimization to train both the appearance and geometry representations, with more details in Sec.~\ref{sec:dense_slam_training}.
The network architecture for the feature network is the same as the depth network, except for the two output branches.
The sizes of channel dimension for the three layers in both the descriptor map and feature map output branches (from hidden to output) are $64$, $64$, and $16$; the output activation functions are both hyperbolic tangent.

\subsection{Differentiable Optimization} \label{sec:dense_slam_diff_optim}
To make the networks learn to master the task of pair-wise image alignment, a differentiable non-linear optimization method is required. 
In this work, we use Levenberg-Marquardt (LM) algorithm as the optimization solver. 
LM is a trust-region algorithm to find a minimum of a function over a space of parameters.
The design is based on Tang~\etal~\cite{tang2018ba}, with modifications to increase memory and computation efficiency.
In the computation graph of network training, all accepted steps in the optimization process are connected, while the decision stage and rejected steps in LM are not involved.
We apply gradient checkpoint technique~\cite{paszke2017automatic} to largely increase the allowed number of accepted steps in the graph.

\subsection{Loss Design}\label{sec:dense_slam_loss_design}
For each iteration, when the LM optimization converges, several outputs before, during, and after the optimization process will be involved in the loss computation for the network training.
The groundtruth data required for training are relative camera pose, camera intrinsics, binary video mask to indicate valid region, dense depth map, and dense 2D scene flow map that can be generated with the data before.
The average and the optimized depth estimate should agree with the groundtruth depth map up to a global scale. 
We do not let the depth network try to predict the correct scale and instead leave it to the optimization during SLAM running because predicting a correct depth scale from a monocular endoscopic image is nearly impossible.
Therefore, a scale-invariant loss is used for this objective. 
With a predicted depth map $\bm{D}\in\mathbb{R}^{1\times H\times W}$, the corresponding groundtruth depth map $\Tilde{\bm{D}}\in\mathbb{R}^{1\times H\times W}$, and the binary video mask $\bm{V}\in\mathbb{R}^{1\times H\times W}$, the loss is defined as
\begin{equation}
    \mathcal{L}_\mathrm{si} = \dfrac{\sum{\bm{D}_\mathrm{ratio}^2}}{\sum{\bm{V}}} + \dfrac{\left(\sum{\bm{D}_\mathrm{ratio}}\right)^2}{\left(\sum{\bm{V}}\right)^2} \quad \text{,}
\end{equation}
where $\bm{D}_\mathrm{ratio} = \log{\left(\bm{V}\bm{D} + \epsilon\right)} - \log{\left(\bm{V}\Tilde{\bm{D}} + \epsilon\right)}$. 
$\epsilon\in\mathbb{R}$ is a small number to prevent logarithm over zero.
To guide the intermediate depth maps during optimization, we additionally use an adversarial loss~\cite{mao2017least}.
In this loss, a discriminator is used to distinguish real samples from fake ones.
The real sample for the GAN will be a color image and the corresponding normalized groundtruth depth map; the fake sample will be the color image and the corresponding normalized depth estimate.
For normalization, these depth maps are divided by their maximum value so that the discriminator judges the fidelity of the sample pair based only on the relative geometry and not on the scale.

For the descriptor map, the RR loss proposed in~\cite{liu2020extremely} is used.
Because a descriptor map is also used for loop closure detection, besides producing good feature matches on images with large scene overlap, having dissimilar descriptions for images with small or no scene overlap is also desired. 
A triplet histogram loss is used to make sure the similarity between histograms of descriptor maps for the source and target images is higher than that for the source and far images. The definitions of these three images are in Sec.~\ref{sec:dense_slam_training}. 
The triplet histogram loss is defined as
\begin{equation}
\begin{split}
    \mathcal{L}_\mathrm{hist} =
    \dfrac{1}{C}\sum_{i\in\{1,\dotsc,C\}}\min
    &\left(
    \dfrac{1}{K}d_\mathrm{EMD}\left(\bm{h}^\mathrm{src}_i,\bm{h}^\mathrm{tgt}_i\right) - \right. \\ 
    &\left. \dfrac{1}{K}d_\mathrm{EMD}\left(\bm{h}^\mathrm{src}_i,\bm{h}^\mathrm{far}_i\right) + \eta_\mathrm{hist}, 0\right) \quad \text{,}
\end{split}
\end{equation}
where $d_\mathrm{EMD}\left(\bm{h}_1,\bm{h}_2\right) = \norm{\mathrm{CDF}\left(\bm{h}_1\right) - \mathrm{CDF}\left(\bm{h}_2\right)}^2$ measures the earth mover's distance between two histograms.
$\mathrm{CDF}$ is the operation to produce cumulative density function (CDF) from a histogram. $\bm{h}^\mathrm{src}_i\in\mathbb{R}^K$ is the soft histogram of elements from source descriptor map $\bm{I}^\mathrm{src}\in\mathbb{R}^{C\times H\times W}$ along the $i$\textsuperscript{th} channel, which is $\bm{I}^\mathrm{src}_i\in\mathbb{R}^{1\times H\times W}$; 
$K$ is the number of bins in each CDF and $C$ is the channel size of the descriptor map;
$\eta_\mathrm{hist}\in\mathbb{R}$ is a constant margin.
To compute the CDF differentiably, we refer to the method in~\cite{aviaharon2020deephist} and describe it in the code repository.

After the optimization process in Sec.~\ref{sec:dense_slam_diff_optim}, the source image should be warped to the target frame with good alignment, using the estimate of status.
Such a warping process can be described with a 2D scene flow. 
Therefore, to guide the learning process to produce better image alignment, another loss is to encourage the similarity between the groundtruth 2D scene flow, and the one estimated from the optimization process.
The flow loss is defines as
\begin{equation}
    \mathcal{L}_\mathrm{flow} = \dfrac{1}{\omega^\mathrm{s\rightarrow t}\sum{\bm{V}}}\sum{\bm{V}\left(\Tilde{\bm{W}}^\mathrm{s\rightarrow t} - \bm{W}^\mathrm{s\rightarrow t}\right)^2}\quad \text{,}
\end{equation}
where $\Tilde{\bm{W}}^\mathrm{s\rightarrow t}\in\mathbb{R}^{2\times H\times W}$ and $\bm{W}^\mathrm{s\rightarrow t}\in\mathbb{R}^{2\times H\times W}$ are the groundtruth and estimated 2D scene flows from source to target frame, respectively.
$\omega^\mathrm{s\rightarrow t}\in\mathbb{R}$ is a normalization factor, defined as $\omega^\mathrm{s\rightarrow t} = \dfrac{1}{2}\sum \bm{V}\left((\Tilde{\bm{W}}^\mathrm{s\rightarrow t})^2 + (\bm{W}^\mathrm{s\rightarrow t})^2\right)$. 
The estimated flow $\bm{W}^\mathrm{s\rightarrow t}$ at 2D location $\bm{x}^\mathrm{src}$ is defined as
\begin{equation}
    \bm{W}^\mathrm{s\rightarrow t}\left(\bm{x}^\mathrm{src}\right) = \pi\left(\bm{p}^\mathrm{s\rightarrow t}\right) - \bm{x}^\mathrm{src}\quad \text{, where}
\end{equation}
\begin{equation}
    \bm{p}^\mathrm{s\rightarrow t} = \bm{T}^\mathrm{tgt}_\mathrm{src}\pi^{-1}\left(\bm{x}^\mathrm{src},\bm{D}^\mathrm{src}\left(\bm{x}^\mathrm{src}\right)\right) \quad \text{.}
\end{equation}
$\bm{p}^\mathrm{s\rightarrow t}\in\mathbb{R}^3$ is the 3D location of the lifted source 2D location $\bm{x}^\mathrm{src}\in\mathbb{R}^2$ in the target coordinate system.
$\pi$ and $\pi^{-1}$ are the project and unproject operation of the camera geometry. 
These two operations are the same for all keyframes because camera intrinsics are assumed to be fixed throughout the video.
$\bm{T}^\mathrm{tgt}_\mathrm{src} = \left(\bm{T}^\mathrm{wld}_\mathrm{tgt}\right)^{-1} \bm{T}^\mathrm{wld}_\mathrm{src}$ is the relative pose between target and source.
$\bm{D}^\mathrm{src}\left(\bm{x}^\mathrm{src}\right)\in\mathbb{R}$ is the depth estimate at 2D location $\bm{x}^\mathrm{src}$ based on the current estimate of depth scale and depth code.
It is defined as $\bm{D}^\mathrm{src}\left(\bm{x}^\mathrm{src}\right) = s^\mathrm{src}\left(\bar{\bm{D}}^\mathrm{src}\left(\bm{x}^\mathrm{src}\right) + \left(\bm{c}^\mathrm{src}\right)^\intercal\hat{\bm{D}}^\mathrm{src}\left(\bm{x}^\mathrm{src}\right)\right)$. The source average depth estimate and depth bases are $\bar{\bm{D}}^\mathrm{src}\in\mathbb{R}^{1\times H\times W}$ and $\hat{\bm{D}}^\mathrm{src}\in\mathbb{R}^{B\times H\times W}$.
the source depth scale, depth code, and camera pose matrix are $s^\mathrm{src}\in\mathbb{R}$, $\bm{c}^\mathrm{src}\in\mathbb{R}^B$, and $\bm{T}^\mathrm{wld}_\mathrm{src}\in\mathrm{SE}\left(3\right)$, respectively.

\begin{figure}
	\centering
	\includegraphics[width=1.0\linewidth]{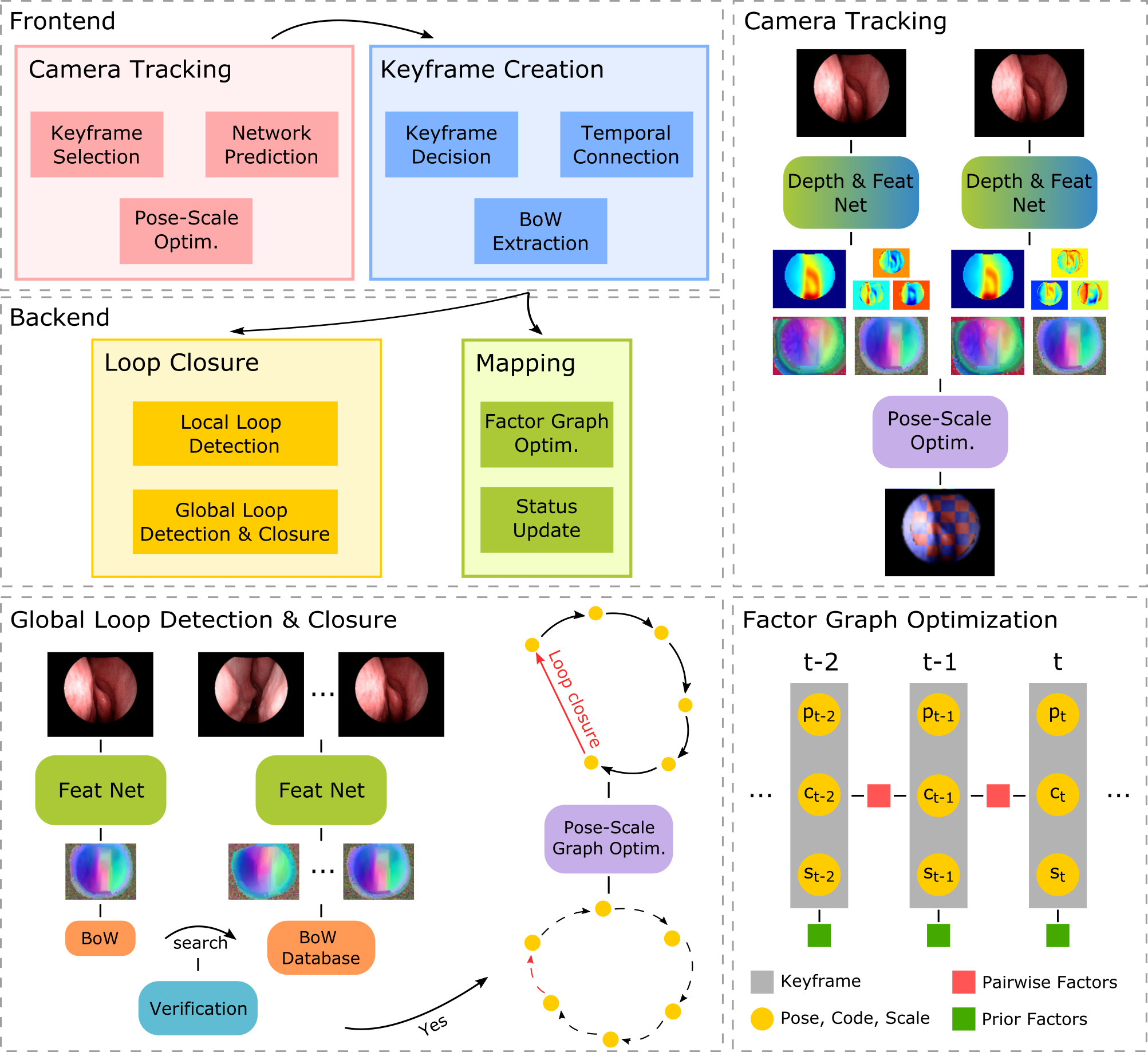}
	\caption[Overall diagram of SLAM system]{\textbf{Overall diagram of SLAM system.}\quad 
	The top left shows the module relationship in our SLAM system.
	The top right demonstrates the network prediction and pose-scale optimization within the \emph{Camera Tracking} module. 
	The bottom left shows the process of global loop detection and closure within the \emph{Loop Closure} module.
	The bottom right demonstrates the optimization in the \emph{Mapping} module, where pair-wise factors between non-adjacent keyframes are not shown.
	More details are described in Sec.~\ref{sec:dense_slam_slam}.
	}
	\label{fig:dense_slam_diagram}
\end{figure}

\subsection{Training Procedure}\label{sec:dense_slam_training}
In each iteration, three images are used for training, which are the source, target, and far images.
Source and target are two images with a large scene overlap, while the far image has a small or no scene overlap with the source.
The network training consists of two stages.
At the first stage, the depth and feature networks are trained separately with the scale-invariant loss and RR loss, respectively.
After both networks are trained to a reasonable state, the training moves to the second stage, where the networks are jointly trained with the scheme below.
The task for training becomes pair-wise image alignment which can be handled well only if networks produce good representations.
The variables that are optimized over are relative camera pose, depth scale, and depth code of the source image.
And the factors involved are pair-wise factors, FM, SMG, and GC, and prior factors, SC and CD, which are introduced in Sec.~\ref{sec:dense_slam_factor_design}.
A random relative camera pose and all-zero depth code are initialized.
The source depth scale is computed to match the scale of the target depth map.

The optimization in Sec.~\ref{sec:dense_slam_diff_optim} is then applied to minimize the objective described by the factors and the networks are updated afterward with the losses described in Sec.~\ref{sec:dense_slam_loss_design}.
A GAN training cycle~\cite{mao2017least} is also involved because of the adversarial loss.

\section{Simultaneous Localization and Mapping}\label{sec:dense_slam_slam}
\subsection{Overview}
The SLAM system modules are organized into frontend and backend threads. 
Frontend consists of \emph{Camera Tracking} and \emph{Keyframe Creation} modules. 
When a new frame comes in, the \emph{Camera Tracking} is used to track it against a reference keyframe.
The \emph{Keyframe Creation} module then handles keyframe creation and temporal keyframe connection.
For each keyframe, a bag-of-words vector is created for global loop detection in the \emph{Loop Closure} module.
Backend threads run \emph{Loop Closure} and \emph{Mapping} modules. 
The \emph{Loop Closure} module constantly detects both local and global connections between all keyframe pairs. 
Whenever a global connection is detected, a lightweight pose-scale graph optimization will be applied to close the loop by adjusting depth scales and camera poses of all keyframes.
The \emph{Mapping} module constantly optimizes all depth codes, depth scales, and camera poses with factors described in Sec.~\ref{sec:dense_slam_factor_design}.
The overall diagram of the SLAM system is shown in Fig.~\ref{fig:dense_slam_diagram}.

\subsection{Factor Design}\label{sec:dense_slam_factor_design}

\textbf{Feature-metric Factor (FM).} \quad
This factor uses the feature map from the feature network as the appearance prior of a frame for reasons in Sec.~\ref{sec:dense_slam_network_arch}.
The feature map is processed to form a Gaussian pyramid with a specified number of levels to increase the convergence basin.
To build a level of the pyramid, the Gaussian smoothing operation with a specified size and sigma, and $2$-time downsampling will be applied to the map in the previous level.
The source feature map pyramid is defined as $\mathcal{\bm{F}^\mathrm{src}} = \{\bm{F}^\mathrm{src}_i | i = 1,\dotsc,L\}$, where L is the number of levels and $\bm{F}^\mathrm{src}_i\in\mathbb{R}^{C\times H/2^{i-1}\times W/2^{i-1}}$ is the feature map at pyramid level $i$;
The objective of this factor is defined as
\begin{equation}
\begin{split}
    \mathcal{L}_\mathrm{fm} = \dfrac{1}{L}\sum_{i=1}^{L}\;\;\dfrac{1}{\vert\Omega_{\mathrm{src},\mathrm{tgt}}\vert}\sum_{\bm{x}^\mathrm{src}\in \Omega_{\mathrm{src},\mathrm{tgt}}}
    &\left\lVert\bm{F}^\mathrm{tgt}_i\left(\pi\left(\bm{p}^\mathrm{s\rightarrow t}\right)\right) - \right. \\
    &\left. \bm{F}^\mathrm{src}_i\left(\bm{x}^\mathrm{src}\right)\right\rVert^2 \quad \text{,}
\end{split}
\end{equation}
where $\Omega_{\mathrm{src},\mathrm{tgt}}$ is the set of source 2D locations that can be projected onto the target mask region given the estimates.

\textbf{Sparse Matched Geometry Factor (SMG).}\quad 
With only FM, the convergence basin is relatively small, which is common for the appearance-warping-based objectives~\cite{yin2018geonet}.
The descriptor map from the feature network can be used to estimate 2D point correspondences between images through feature matching.
This enables the objective to have global convergence characteristics.
Because in this work, each keyframe has a depth estimate, the 2D correspondences can be replaced with 3D ones.
Compared with 2D, the 3D ones should contain fewer outliers because 3D point cloud alignment~\cite{Yang2020teaser} is used to remove outliers, which has less ambiguity than the common 2D filtering method based on epipolar geometry.
The definition of this factor is: 
\begin{equation}
\begin{split}
    \mathcal{L}_\mathrm{smg} = \dfrac{1}{\vert\mathcal{M}\vert}
    &\sum_{\left(\bm{x}^\mathrm{src}, \bm{x}^\mathrm{tgt}\right)\in \mathcal{M}}\rho_\mathrm{fair}\left(\lVert\bm{p}^\mathrm{s\rightarrow t} - \right.\\
    &\left. \pi^{-1}\left(\bm{x}^\mathrm{tgt},\bm{D}^\mathrm{tgt}\left(\bm{x}^\mathrm{tgt}\right)\right)\rVert^2; \delta^\mathrm{src}_\mathrm{smg}\right)\quad \text{,}
\end{split}
\end{equation}
where $\mathcal{M}$ is a set of feature matches consisting of pairs of 2D locations $\left(\bm{x}^\mathrm{src},\bm{x}^\mathrm{tgt}\right)\in\mathbb{R}^2\times\mathbb{R}^2$, and $\delta^\mathrm{src}_\mathrm{smg} = \dfrac{\sigma_\mathrm{smg}}{\vert\Omega^\mathrm{src}\vert}\sum_{\bm{x}\in\Omega^\mathrm{src}}\bar{\bm{D}}^\mathrm{src}\left(\bm{x}\right)$, which is the mean value of the source average depth estimate multiplying a constant factor $\sigma_\mathrm{smg}\in\mathbb{R}$.
The outlier-robust "Fair" loss~\cite{bosse2016robust} is used, which is defined as $\rho_{\mathrm{fair}}\left(a;b\right) = 2(\sqrt{a/b} - \ln{(1 + \sqrt{a/b})})$.

\textbf{Reprojection Factor (RP).}\quad
This factor behaves similarly to SMG except that the objective is changed from minimizing the average distance of 3D point sets to that of the corresponding projected 2D locations.
The factor is defined as: 
\begin{equation}
\begin{split}
    &\mathcal{L}_\mathrm{rp} = 
    \dfrac{1}{\vert\mathcal{M}\vert}\sum_{\left(\bm{x}^\mathrm{src}, \bm{x}^\mathrm{tgt}\right)\in \mathcal{M}} \\
    &\rho_{\mathrm{fair}}\left(\lVert\pi\left(\bm{T}^\mathrm{tgt}_\mathrm{src}\pi^{-1}\left(\bm{x}^\mathrm{src},\bm{D}^\mathrm{src}\left(\bm{x}^\mathrm{src}\right)\right)\right) - \bm{x}^\mathrm{tgt}\rVert^2; \sigma_\mathrm{rp} W^2\right)\quad \text{,}
\end{split}
\end{equation}
where $\sigma_\mathrm{rp}\in\mathbb{R}$ is a multiplying factor and $W$ is the width of the involved depth map.

\textbf{Geometric Consistency Factor (GC).}\quad
This factor enforces geometric consistency by encouraging the source depth estimate transformed to the target coordinate to have consistent values as the target depth estimate.
The factor is defined as: 
\begin{equation}
\begin{split}
\mathcal{L}_\mathrm{gc} = \dfrac{1}{\vert\mathcal{M}\vert}\sum_{\left(\bm{x}^\mathrm{src}, \bm{x}^\mathrm{tgt}\right)\in \mathcal{M}}
&\rho_{\mathrm{cauchy}}\left(\lVert z^\mathrm{s\rightarrow t} - \right. \\
&\left. \bm{D}^\mathrm{tgt}\left(\pi\left(\bm{p}^\mathrm{s\rightarrow t}\right)\right)\rVert^2; \delta_\mathrm{gc}^\mathrm{src}\right)\quad \text{,}
\end{split}
\end{equation}
where $z^\mathrm{s\rightarrow t}$ is the z-axis component of $\bm{p}^\mathrm{s\rightarrow t}$;
$\delta_\mathrm{gc}^\mathrm{src}$ is the same as $\delta_\mathrm{smg}^\mathrm{src}$, except that $\sigma_\mathrm{gc}$ is used instead of $\sigma_\mathrm{smg}$.
Cauchy loss~\cite{bosse2016robust} is used to increase the robustness of this factor, which is defined as $\rho_\mathrm{cauchy}\left(a;b\right) = \ln{\left(1 + a/b\right)}$.

\textbf{Relative Pose Scale Factor (RPS).}\quad 
This factor is used in the graph optimization for the global loop closure in Sec.~\ref{sec:dense_slam_loop_closure}.
The error value for the pair-wise factors above will not change if the depth scales and relative camera pose are scaled jointly.
During a global loop closure, all frame pairs except the newly detected global loop should have reasonably variable estimates.
Therefore, this factor is to keep variables in the previous links unchanged up to a global scale and encourage the new global link to reach the goal.
The factor is defined as follows: 
\begin{equation}
\begin{split}
    \mathcal{L}_\mathrm{rps} = 
    &\norm{\dfrac{\bm{t}^\mathrm{tgt}_\mathrm{src}}{s^\mathrm{src}} - \dfrac{\Tilde{\bm{t}}^\mathrm{\;tgt}_\mathrm{\;src}}{\Tilde{s}^\mathrm{\;src}}}^2 + 
    \omega_\mathrm{rot}\norm{\log{\left(\bm{R}^\mathrm{tgt}_\mathrm{src}\right)} - \log{\left(\Tilde{\bm{R}}^\mathrm{tgt}_\mathrm{src}\right)}}^2 + \\
    &\omega_\mathrm{scl}\left(\log{\left(\dfrac{s^\mathrm{tgt}}{s^\mathrm{src}}\right)} - \log{\left(\dfrac{\Tilde{s}^\mathrm{\;tgt}}{\Tilde{s}^\mathrm{\;src}}\right)}\right)^2 \quad \text{,}
\end{split}
\end{equation}
where $\bm{t}^\mathrm{tgt}_\mathrm{src}\in\mathbb{R}^3$ and $\bm{R}^\mathrm{tgt}_\mathrm{src}\in\mathrm{SO}\left(3\right)$ are the translation and rotation components of the relative pose $\bm{T}^\mathrm{tgt}_\mathrm{src}$ described above, respectively.
Note that the logarithm operation on the rotation components is the matrix logarithm of $\mathrm{SO}\left(3\right)$.
$\omega_\mathrm{rot}\in\mathbb{R}$ and $\omega_\mathrm{scl}\in\mathbb{R}$ are the weights for the rotation and scale components of this factor, respectively.
In this equation and the ones below, every symbol with $\Tilde{}$ on top represents the target counterpart of the one without it.

\textbf{Code Factor (CD).}\quad 
This is used to keep the depth code of a keyframe within a reasonable range. 
It is defined as
\begin{equation}
\mathcal{L}_\mathrm{code} = \dfrac{1}{B}\norm{\bm{c}^\mathrm{src} - \Tilde{\bm{c}}^\mathrm{\;src}}^2 \quad \text{.}
\end{equation}

\textbf{Scale Factor (SC).}\quad 
This is to make the depth scale of a keyframe close to the goal.
It is defined as
\begin{equation}
\mathcal{L}_\mathrm{scale} = \left(\log{\left(s^\mathrm{src}\right)} - \log{\left(\Tilde{s}^\mathrm{\;src}\right)}\right)^2 \quad \text{.}
\end{equation}

\textbf{Pose Factor (PS).}\quad 
It is used in the first keyframe to anchor the trajectory of the entire graph, which is defined as
\begin{equation}
\begin{split}
\mathcal{L}_\mathrm{pose} = 
&\norm{\bm{p}^\mathrm{wld}_\mathrm{src} - \Tilde{\bm{p}}^\mathrm{\;wld}_\mathrm{\;src}}^2 + \\ &\omega_\mathrm{r}\norm{\log{\left(\bm{R}^\mathrm{wld}_\mathrm{src}\right)} - \log{\left(\Tilde{\bm{R}}^\mathrm{wld}_\mathrm{src}\right)}}^2 \quad \text{,}
\end{split}
\end{equation}
where $\omega_\mathrm{r}\in\mathbb{R}$ is the weight of the rotation component.

\subsection{Module Design}

\textbf{Camera Tracking.}\quad
This module is used to track a new frame against a reference keyframe.
The reference is the spatially closest one against the last frame, which is verified based on appearance similarity.
Camera tracking is solved with LM optimization over the relative camera pose, $\bm{T}^\mathrm{tgt}_\mathrm{src}$, between the new frame and the reference, where factors FM and RP are involved.
The termination of optimization is based on several criteria, which are the maximum number of iterations, parameter update ratio threshold, and gradient threshold.
Once the optimization finishes, the pose of the new frame can then be computed correspondingly.

\textbf{Keyframe Creation.}\quad
For every tracked new frame, this module first determines if a new keyframe is needed. 
Because the scale of the entire graph is ambiguous due to the scale ambiguity of monocular depth estimation, no absolute distance threshold can be relied on.
Instead, we use a set of more intuitive criteria that directly relate to the information gain of a new frame, which are scene overlap, feature match inlier ratio, and the average magnitude of 2D scene flow.
Scene overlap measures the overlap between two frames and reflects how much new region is observed from a new frame.
Feature match inlier ratio is the ratio of inlier matches over all the feature match candidates.
This reflects how dissimilar the two frames are in terms of appearance, which may be due to a small region overlap, a dramatic texture change, \etc.
The average magnitude of 2D scene flow measures how much movement the content of a frame has. 
This is to track the camera movement of keyframes more continuously and to produce more consistent descriptors and feature maps between keyframes.
For each keyframe, a bag-of-words vector is computed from the descriptor map and added to a database for global loop indexing in Sec.~\ref{sec:dense_slam_loop_closure}.
Connections consisting of keyframes within a temporal range will be added to the new keyframe.
At least one keyframe will be connected to the new one and extra ones, up to a specified number, will be added only if the appearance is similar to the new keyframe.
The pair-wise factors involved in the keyframe connections are FM and GC.
For the first keyframe, prior factors, CD, SC, and PS, are integrated into the factor graph and only CD will be included for the other keyframes.

\begin{table*}[!t]
\renewcommand{\arraystretch}{1.0}
\caption[Cross-subject evaluation on SLAM systems]{\textbf{Cross-subject evaluation on SLAM systems.}}
\label{tab:dense_slam_cross_evaluation_overall}
\centering
\resizebox{1.0\linewidth}{!}{
\begin{tabular}{lllllllllll}
\hline
\makecell[l]{Metrics \\/ Methods} & \makecell[l]{$\mathrm{ATE}_\mathrm{trans}$ \\ $\mathrm{(mm)}$} & \makecell[l]{$\mathrm{ATE}_\mathrm{rot}$ \\ $\mathrm{(\degree)}$} & \makecell[l]{$\mathrm{RPE}_\mathrm{trans}$ \\ $\mathrm{(mm)}$} & \makecell[l]{$\mathrm{RPE}_\mathrm{rot}$ \\ $\mathrm{(\degree)}$} & $\mathrm{ARD}_\mathrm{traj}$ & $\mathrm{ARD}_\mathrm{frame}$ & \makecell[l]{$\mathrm{Threshold}_\mathrm{traj}$ \\ $(\theta=1.25)$} & \makecell[l]{$\mathrm{Threshold}_\mathrm{frame}$ \\ $(\theta=1.25)$} & \makecell[l]{$\mathrm{Threshold}_\mathrm{traj}$ \\ $(\theta=1.25^2)$} & \makecell[l]{$\mathrm{Threshold}_\mathrm{frame}$ \\ $(\theta=1.25^2)$} \\ \hline
Ours & $\bm{1.6\pm1.4}$ & $\bm{22.2\pm15.1}$ & $\bm{1.5\pm0.6}$ & $\bm{5.5\pm2.4}$ & $\bm{0.36\pm0.16}$ & $\bm{0.17\pm0.03}$ & $\bm{0.42\pm0.17}$ & $\bm{0.73\pm0.08}$ & $\bm{0.74\pm0.21}$ & $\bm{0.95\pm0.04}$ \\
ORB-SLAM v3~\cite{ORBSLAM32020} & $4.7\pm4.2^\text{***}$ & $62.5\pm55.5^\text{***}$ & $3.5\pm2.5^\text{***}$ & $6.3\pm3.6$ & $1.76\pm1.49^\text{***}$ & $24.27\pm42.07^\text{**}$ & $0.17\pm0.18^\text{***}$ & $0.37\pm0.13^\text{***}$ & $0.31\pm0.25^\text{***}$ & $0.56\pm0.15^\text{***}$ \\ \hline
\end{tabular}
}
\end{table*}

\textbf{Mapping.}\quad
The mapping is constantly running at the backend, where the framework of optimization is ISAM2~\cite{Kaess2012isam2}.
The factor graph consisting of pair-wise and prior factors from all keyframes is optimized in this module and Fig.~\ref{fig:dense_slam_diagram} shows an example of such a graph.
The variables jointly optimized are camera poses, depth scales, and depth codes of all keyframes.

\textbf{Loop Closure.}\quad \label{sec:dense_slam_loop_closure}
This module constantly tries to search for local or global loop connections through all keyframes and handles the closure correspondingly.
For local loop detection, the keyframes within a specified temporal range for each searched keyframe are considered.
A verification based on filtering, appearance, and geometric, which is described with more details in the code repository, is applied to select the best local loop candidate.
The selected local connection is then linked with pair-wise factors same as the temporal connections.
Another part of this module is global loop connection and closure, as shown in Fig.~\ref{fig:dense_slam_diagram}.
Global loop detection searches for keyframe pairs whose interval is beyond a specified temporal range and first applies appearance verification.
The descriptor map from the feature network describes the appearance distinctively and those from the training set are used to build a bag-of-words place recognition model~\cite{Galvez2012DBoW2}.
When a global loop connection is searched for a query keyframe, the database will be searched through with the query bag-of-words vector.
A specified number of keyframes that are the most similar to the query keyframe in terms of bag-of-words description will be selected as candidates.
Then the appearance and geometric verification, which is described in the code repository, is applied to select global loop candidates.
The drifting error for the global connection is often large.
Therefore, it is slow to rely on the full graph optimization in the \emph{Mapping} module to close the gap.
To this end, we design a lightweight pose-scale graph optimization for the global loop closure, where all camera poses and depth scales are optimized jointly.
In this graph, a set of lightweight factors are used.
For the new global loop pair, SC and RPS are used, where the target camera poses and depth scales come from the geometric verification above; 
For all other keyframe connections, RPS is used, where the values of the current estimates are used as the target in the factors.
The graph optimization terminates if the maximum number of iterations is reached or the number of updates with no relinearization reaches a threshold.

\section{Experiments}
\subsection{Cross-Subject Evaluation} \label{sec:dense_slam_cross_subject}
Please refer to the code repository for the experiment setup in terms of parameter setting.
For all studies in this work, the metrics used for camera trajectory evaluation are Absolute Trajectory Error (ATE) and Relative Pose Error (RPE)~\cite{zhang2018trajectory}.
Note that only the frames that are treated as keyframes by the SLAM system will be evaluated in terms of both trajectory error and depth error.
Therefore, synchronization needs to be done to first associate the trajectory estimate with the groundtruth one.
The trajectory estimate will then be spatially aligned with the groundtruth trajectory, where a similarity transform is estimated~\cite{zhang2018trajectory}.
To evaluate depth estimates, Absolute Relative Difference (ARD) and Threshold~\cite{yin2018geonet} are used.
Before computing metrics, different pre-processing is applied for two sets of metrics, which are $\mathrm{ARD}_\mathrm{traj}$ and $\mathrm{Threshold}_\mathrm{traj}$, and $\mathrm{ARD}_\mathrm{frame}$ and $\mathrm{Threshold}_\mathrm{frame}$.
For the first set, the estimated depth per keyframe is scaled with the scale component in the similarity transform obtained from the trajectory alignment above.
For the other set, each depth estimate is scaled with the median value of ratios between the corresponding groundtruth one and the estimate.
Please find definitions of these metrics in the code repository.
To evaluate the performance of the SLAM system on endoscopic videos from unseen subjects, we run a cross-validation study.
Four models are trained with different train/test splits on $11$ subjects in total, where each test split has $3$ subjects and the rest are used for training.
For evaluation, the proposed SLAM is run on all sequences, with runtime performance of around $5.5$ FPS, and generates estimates of camera poses and dense depth maps for all keyframes.
Note that the value of each metric is averaged over all the sequences from all subjects, where each subset of the sequences is evaluated with the corresponding trained model so that all the sequences are unseen during training.
We also compare against ORB-SLAM3~\cite{ORBSLAM32020}.
We adjust its parameters so that more keypoint candidates can be detected per frame.
We conduct the paired t-test analysis between results from ORB-SLAMv3 and the proposed system.
Evaluation results are shown in Table~\ref{tab:dense_slam_cross_evaluation_overall}, where the values with $^\text{***}$, $^\text{**}$, and $^\text{*}$ stand for p-value smaller than 0.001, 0.01, and 0.05, respectively.
The proposed system outperforms ORB-SLAM3 on all metrics with statistical significance except for the $\mathrm{RPE}_\mathrm{rot}$ where $p = 0.12$.

\subsection{Ablation Study}
We evaluate the contributions of several SLAM components by disabling some in different runs.
The components for ablation are FM in the \emph{Camera Tracking} and \emph{Mapping} modules (FMT and FMM), RP in the \emph{Camera Tracking} module (RPT), local loop detection in the \emph{Loop Closure} module (Local), and global loop detection and closure in the \emph{Loop Closure} module (Global).
All metrics described in Sec.~\ref{sec:dense_slam_cross_subject} are evaluated and results are provided in the code repository.
The overall observation is, FM has a large impact on both trajectory and trajectory-scaled depth metrics;
RP mainly affects trajectory metrics;
the \emph{Loop Closure} module mainly affects the trajectory metrics $\mathrm{ATE}_\mathrm{trans}$ and $\mathrm{ATE}_\mathrm{rot}$.

\subsection{Evaluation with CT}\label{sec:dense_slam_ct}
This study uses the average residual error between the registered surface reconstruction and the corresponding CT model as the evaluation metric.
Before computing the residual error, the depth fusion method in~\cite{Curless1996tsdf} is first applied to obtain a surface reconstruction from SLAM output.
Then a point cloud registration algorithm based on~\cite{billings2015generalized} is applied between the surface reconstruction and the CT surface model, where a similarity transform is estimated.
Lastly, the residual error is computed between the registered surface reconstruction and the CT surface model.
In this study, we evaluate the accuracy of surface reconstructions from the videos of $4$ cadavers, where for each subject, the metrics of all the sequences are averaged over to report here.
The average residual errors for subject $7$, $9$, $10$, and $11$ are $0.83$, $0.88$, $0.78$, and $0.86\;\mathrm{mm}$, respectively.

\section{Conclusion}
In this work, we propose a SLAM system, integrated with learning-based appearance and optimizable geometric priors, that can track the endoscope and reconstruct dense geometry of the anatomy from a monocular endoscopic video stream.
An effective end-to-end training pipeline is developed to learn such priors by explicitly mastering the task of pair-wise image alignment.
Based on the experiments, the system is shown to be robust to texture-scarce and illumination-varying scenarios and generalizable to unseen endoscopes and patients.
To serve as a brief discussion, the accuracy of the proposed SLAM system depends on the generalizability of networks, and thus a representative collection of data for network training is important.
Currently, the system cannot recover from a spurious global loop connection, which might be enabled with~\cite{Cadena2016Past}, and therefore the global loop detection criteria need to be strict to keep the false positive rate to zero.
The current system is designed for static scenes, though having additional variables to model deformation (\eg,~deformation-spline~\cite{Kopf2021Robust}) could also make the system suitable for deformable environments.

\bibliographystyle{IEEEtran}
\bibliography{Bibliography}

\end{document}


\title{Supplementary Material -- SAGE: SLAM with Appearance and Geometry Prior for Endoscopy}
\author{\IEEEauthorblockN{Xingtong~Liu, Zhaoshuo~Li, Masaru~Ishii, Gregory~D.~Hager,~\IEEEmembership{Fellow,~IEEE,} Russell~H.~Taylor,~\IEEEmembership{Life Fellow,~IEEE,} and Mathias~Unberath}%
    \thanks{Xingtong~Liu, Zhaoshuo~Li, Gregory~D.~Hager, Russell~H.~Taylor, and Mathias~Unberath are with the Computer Science Department, Johns Hopkins University, Baltimore, MD 21287 USA (e-mail: xliu89@jh.edu; zli122@jhu.edu; hager@cs.jhu.edu; rht@jhu.edu; unberath@jhu.edu).}
    \thanks{Masaru Ishii is with Johns Hopkins Medical Institutions, Baltimore, MD 21224 USA (e-mail: mishii3@jhmi.edu).}
}

\maketitle

\section{Network Architecture}

\begin{figure}
	\centering
	\includegraphics[width=0.9\linewidth]{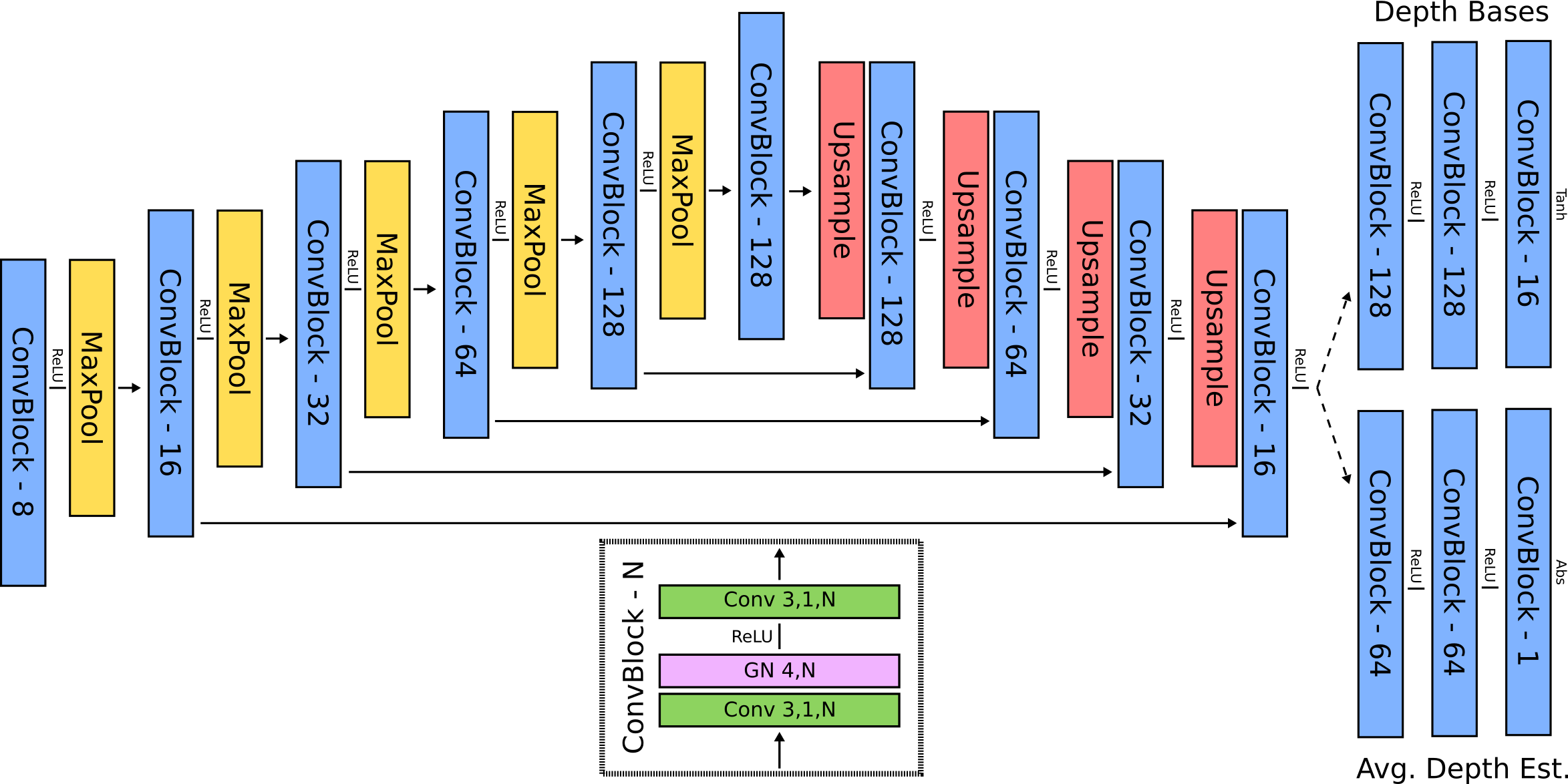}
	\caption[Network architecture for optimizable depth estimation]{\textbf{Network architecture for optimizable depth estimation.}\quad 
	Each ConvBlock consists of two partial convolution layers with kernel size as 3 and stride as 1, one group normalization layer with a group size of 4, and one ReLU activation, which are arranged in the way as the figure above. 
	The number after the ConvBlock means the size of the output channel dimension. 
	Two output branches exist in the network for the average depth estimate and the depth bases.
	Hyperbolic tangent and absolute functions are used as output activation in these branches.
	}
	\label{fig:dense_slam_depth_net}
\end{figure}

\begin{figure}[h]
	\centering
	\includegraphics[width=0.9\linewidth]{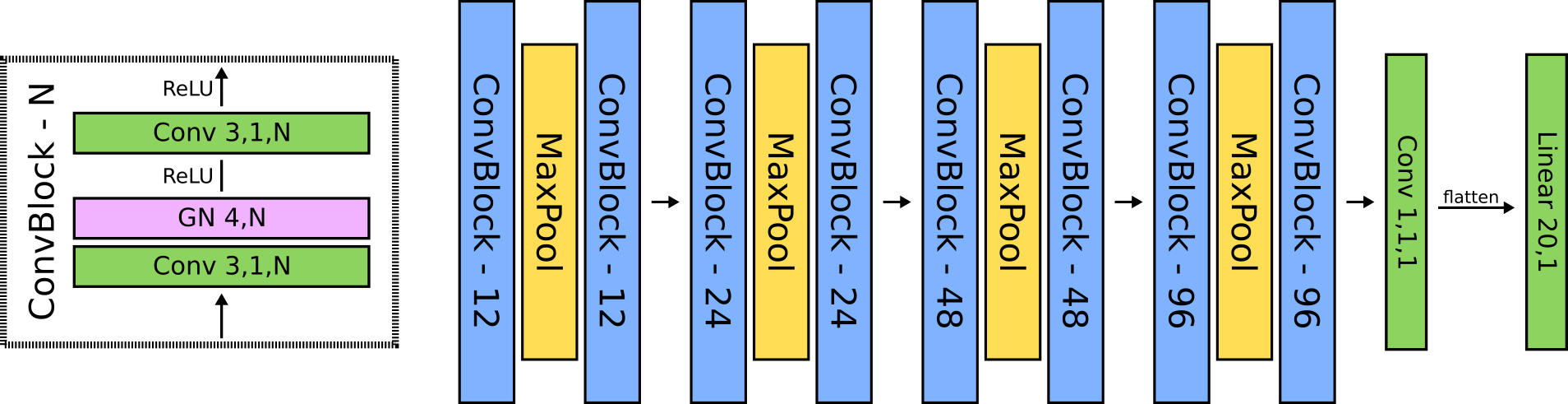}
	\caption[Network architecture of the discriminator for depth estimation learning]{\textbf{Network architecture of discriminator for depth estimation learning.}\quad 
	The input is the RGB image and the normalized depth map, concatenated along the channel dimension, with a resolution of $64\times80$.
	Each ConvBlock consists of two normal convolution layers with kernel size as 3 and stride as 1, one group normalization layer with a group size of 4, and two ReLU activation layers, which are arranged in the way as the figure above. 
	The number after the ConvBlock means the size of the output channel dimension. 
	The final convolution layer, with kernel size as 1, stride as 1, and output channel size as 1, and linear layer, with input channel size as 20 and output channel size as 1, converts the feature map to a scalar value used to indicate the predicted validity of the input sample.
	Note that before being fed to the final linear layer, the output map from the final convolution layer is first flattened along the sample-wise dimensions.
	}
	\label{fig:dense_slam_disc_net}
\end{figure}

The network architecture of the depth network is shown in Fig.~\ref{fig:dense_slam_depth_net} and that of the discriminator used for depth training is shown in Fig.~\ref{fig:dense_slam_disc_net}.

\section{Soft Cumulative Density Function (CDF) Computation}
The value of $k$\textsuperscript{th} bin in the histogram $\bm{h}^\mathrm{src}_i$ can be written as follows
\begin{equation}
\begin{split}
    \bm{h}^\mathrm{src}_i\left(k\right) = 
    \dfrac{1}{\vert \Omega^\mathrm{src}\vert}
    \sum_{\bm{x}\in\Omega^\mathrm{src}}
    &\left(
    \sigma\left(\dfrac{\bm{I}^\mathrm{src}_i\left(\bm{x}\right) - \mu_k + 1/K}{\beta}
    \right) - \right. \\
    &\left. \sigma\left(\dfrac{\bm{I}^\mathrm{src}_i\left(\bm{x}\right) - \mu_k - 1/K}{\beta}\right)
    \right)
    \quad \text{,}
\end{split}
\end{equation}
where the center value of $k$\textsuperscript{th} bin is $\mu_k = -1 + \left(2k + 1\right) / K \in\mathbb{R}$;
the kernel function is $\sigma\left(a\right) = 1 / \left(1 + \mathrm{e}^{-a}\right)$.
The values used in $\mu_k$ are because the descriptor map has a value range of $\left(-1, 1\right)$.
$\Omega^\mathrm{src}$ consists of all 2D locations within the source video mask;
$\beta\in\mathbb{R}$ is a bandwidth parameter.
All other notations are defined in the main paper.
The histograms for target and source images are the same as above except the corresponding descriptor maps are used for calculation instead of the source one.

\section{Verification for Loop Detection}
For the local loop detection, because the temporal window is set to be small, the trajectory drifting error will not be large.
The camera pose of each keyframe can thus still be roughly relied on for filtering candidates.
For this reason, the spatial distance between keyframe pairs is first used.
For the verification, the query keyframe and the closest one within its temporal connections are used as the reference pair.
If the spatial distance between the candidate pair is smaller than that between the reference pair multiplying a constant factor, the pair will be kept.
For pairs being kept after distance filtering, the appearance verification will be run, where the feature match inlier ratio is computed.
The candidate pair will be kept if the inlier ratio is larger than that of the reference pair multiplying a constant factor and a specified constant inlier ratio.
Lastly, a geometric verification is applied, where a pair-wise optimization similar to \emph{Camera Tracking} is run.
The difference in terms of factors is that SMG is used in place of RP.
The local connection will only be accepted if the overlap ratio and flow magnitude, computed in the geometric verification, are larger and smaller than those of the reference pair multiplying a constant factor, respectively.
After verification, only the best candidate, in terms of overlap ratio and flow magnitude, will be used to build the local connection.

Global loop detection searches for keyframe pairs whose interval is beyond a specified temporal range and uses the appearance for the initial candidate selection
Whenever a keyframe is created, the bag-of-words descriptor will be added to a database.
When a global loop connection is searched for a query keyframe, the database will be searched through with the extracted bag-of-words descriptor.
A specified number of keyframes that are similar to the query keyframe in terms of bag-of-words descriptor will be selected as candidates.
The candidates are then filtered so that the similarities between the query keyframe and candidates are larger than the one between the reference pair multiplying a specified constant factor.
Candidates should not be temporally close to the query keyframe, opposite to the local loop connection.
After that, the same verification as the local loop detection is used to verify the global loop candidates.
The verified candidates are ranked based on feature match inlier ratio and, from high to low, each candidate that is temporally far enough from the selected candidates is added to avoid connection redundancy.

\section{Experiment Setup}
The endoscopic videos used in the experiments were acquired from seven consenting patients and four cadavers under an IRB-approved protocol.
The anatomy captured in the videos is the nasal cavity.
The total time duration of videos is around $40$ minutes.
The input images to both networks are $8$-time spatially downsampled, resulting in a resolution of $128\times 160$; the output maps of both networks have a resolution of $64\times 80$.
Note that the binary masks with the same resolution are also fed, together with images, into the networks to exclude contributions of invalid pixels.
SGD optimizer with cyclic learning rate scheduler~\cite{smith2017cyclical} is used for network training, where the learning rate range is $[1.0\mathrm{e}^{-4}, 5.0\mathrm{e}^{-4}]$.
Full-range rotation augmentation is used for input images to the networks during training.
The first stage of training lasts for $40$ epochs and the second stage lasts until the loss curves plateau, where each epoch consists of 300 iterations with the batch size of 1.
Image pairs are selected so that the groundtruth ratio of scene overlap is larger than $0.6$; the initialized relative pose is randomized so that the initial ratio of scene overlap is larger than $0.4$. 
The weights for scale-invariant loss, RR loss, flow loss, histogram loss, generator adversarial loss, and discriminator adversarial loss are $20.0$, $4.0$, $10.0$, $4.0$, $1.0$, and $1.0$.
In terms of the hyperparameters related to loss design, $\epsilon$ is $1.0\mathrm{e}^{-4}$; 
$\eta_\mathrm{hist}$ is $0.3$; 
$\beta$ is $\dfrac{4}{5K}$; 
$K$ is $100$; 
$C$ is $16$; 
$H$ is $64$; 
$W$ is $80$; 

For SLAM system running, in cases where post-operative processing in a SLAM system is allowed, the \emph{Mapping} and \emph{Loop Closure} modules can be run for an additional amount of time after all frames have been tracked.
The \emph{Mapping} module will continue refining the full factor graph.
The maximum number of iterations and consecutive no-relinearization iterations are $20$ and $5$, respectively.
In the meantime, the \emph{Loop Closure} module will search for loop pairs for the query keyframes that have not been processed before.
When the \emph{Mapping} module finishes, the entire system run will end.
For the other settings in terms of the SLAM system running, please refer to the supplementary material.

In terms of the hyperparameters of the differentiable LM optimization, damp value range is $[1.0\mathrm{e}^{-6}, 1.0\mathrm{e}^{-2}]$, with $1.0\mathrm{e}^{-4}$ as the initial value.
The increasing and decreasing multiplier of the damp value is $11.0$ and $9.0$, respectively.
LM optimization terminates when one of the three below is met: 1) number of iterations reaching $40$, 2) maximum gradient smaller than $1.0\mathrm{e}^{-4}$, 3) maximum parameter increment ratio smaller than $1.0\mathrm{e}^{-2}$.
Factors involved have the same parameter setting as the SLAM system, which will be described below.

Below are the hyperparameters of the SLAM system.
For the \emph{Camera Tracking} module, the multiplying factor used for the reference keyframe selection is $0.6$;
the maximum number of iterations in the optimization is $40$;
the damp value range is $[1.0\mathrm{e}^{-6}, 1.0\mathrm{e}^{-2}]$, with $1.0\mathrm{e}^{-4}$ as the initial value;
the increasing and decreasing multipliers are $100.0$ and $10.0$, respectively;
the jacobian matrix recompute condition is when the error update between steps is larger than $1.0\mathrm{e}^{-2}$ of the current error.
As for factors in the \emph{Camera Tracking} module, settings are as follows.
In FM, all samples within the video mask are used for computation;
the weights for all $4$ pyramid levels (from high resolution to low one) are $10.0$, $9.0$, $8.0$, and $7.0$.
In RP, the factor weight and $\sigma_\mathrm{rp}$ are $0.1$ and $0.03$, respectively.
In SMG, the factor weight and $\sigma_\mathrm{smg}$ are $0.1$ and $0.1$, respectively;
the number of feature match candidates before filtering is $256$;
in terms of the Teaser++ filtering, the maximum clique time limit, rotation maximum iterations, rotation graph, inlier selection mode, and noise bound multiplier are $50\mathrm{ms}$, $20$, chain mode, no inlier selection, and $2.0$, respectively;
Other parameters of Teaser++ are set to the default ones.

For the \emph{Keyframe Creation} module, settings are as below.
The maximum ratios of scene overlap in terms of the area and the number of point inliers within the video mask for a new keyframe are $0.8$ and $0.9$, respectively;
the maximum feature match inlier ratio is $0.4$;
the minimum average magnitude of 2D flow is $0.08$ of the image width.
For the temporal connection building in the \emph{Keyframe Creation} module, the maximum number of temporal connections per keyframe is $3$;
the minimum feature match inlier ratio to connect a previous keyframe is $0.7$.

For the \emph{Loop Closure} module, settings are shown as follows.
For the local loop detection, the temporal window for searching is $9$;
the rotation and translation weights to compute pose distance for candidate filtering are both set to $1.0$;
the spatial distance multiplier for candidate filtering is $5.0$;
the metric multiplier for verification is $0.7$;
the minimum constant inlier ratio for verification is $0.2$, which is the same in global loop detection;
the minimum ratios of scene overlap for verification in terms of the area and the number of point inliers within the video mask are $0.5$ and $0.5$, respectively.

Regarding the global loop detection, only keyframes that are at least $10$ keyframes away are considered;
the multiplier of description similarity for verification is $0.7$;
the metric multiplier for verification is $0.7$;
a global loop candidate will be selected if it is at least $10$ keyframes away from the ones already selected in a single global loop closure process.
In the pose-scale graph optimization for loop closure, the weights of RPS for non-global and global connections are $1.0$ and $5.0$, respectively;
within this factor, the weights of rotation and scale component, which are $\omega_\mathrm{rot}$ and $\omega_\mathrm{scl}$, are $5.0$ and $0.5$, respectively;
the weight of SC within the loop closure optimization is $10.0$;
the number of maximum iterations of such optimization is $200$;
the number of maximum iterations with no relinearization is $5$;
the relinearization thresholds for pose and scale are $3.0\mathrm{e}^{-3}$ and $1.0\mathrm{e}^{-2}$.

For the \emph{Mapping} module, settings are as follows.
In terms of hyperparameters of factors used in the full factor graph,
the weights for PS and SC of the first keyframe are $1.0\mathrm{e}^4$, which are used to anchor the graph in terms of camera pose and depth scale;
The FM and GC use all samples within the video mask for computation;
FM has the same weight as the one in camera tracking;
GC has the factor weight of $0.1$ and $\sigma_\mathrm{gc}$ as $0.03$;
the weight of CD is $1.0\mathrm{e}^{-4}$.
In terms of the hyperparameters in factor graph optimization algorithm ISAM2~\cite{Kaess2012isam2}, the relinearization thresholds for camera poses, depth scales, and depth codes are $1.0\mathrm{e}^{-3}$, $1.0\mathrm{e}^{-3}$, and $1.0\mathrm{e}^{-2}$, respectively;
partial relinearization check and relinearization skipping are not used;
Other parameters in ISAM2 are set to the default ones.

In cases where post-operative processing in a SLAM system is allowed, the \emph{Mapping} and \emph{Loop Closure} modules can be run for an additional amount of time after all frames have been tracked.
The \emph{Mapping} module will continue refining the full factor graph.
The maximum number of iterations and consecutive no-relinearization iterations are $20$ and $5$, respectively.
In the meantime, the \emph{Loop Closure} module will search for loop pairs for the query keyframes that have not been processed before.
When the \emph{Mapping} module finishes, the entire system run will end.

\begin{table*}[!]
\renewcommand{\arraystretch}{1.0}
\caption[Cross-subject evaluation on SLAM systems]{
\textbf{Cross-subject evaluation on SLAM systems.} 
Note that $\sim$ is used as the name abbreviation of the comparison method ORB-SLAM3.}
\label{tab:dense_slam_cross_evaluation}
\centering
\resizebox{0.9\linewidth}{!}{
\begin{tabular}{@{}lcccccccc@{}}
\toprule
 Subjects & \multicolumn{2}{c}{$\{1, 2, 3\}$} & \multicolumn{2}{c}{$\{4, 5, 6\}$} & \multicolumn{2}{c}{$\{7, 8, 11\}$} & \multicolumn{2}{c}{$\{8, 9, 10\}$} \\ \midrule
 \makecell[l]{Methods \\/ Metrics} &   Ours    &   ORB-SLAM3~\cite{ORBSLAM32020}   &     Ours      &    $\sim$      &      Ours     &   $\sim$      &     Ours      &     $\sim$     \\ \midrule
 $\mathrm{ATE}_\mathrm{trans} \mathrm{(mm)}$ &    $\bm{1.4\pm1.0}$      &    $3.8\pm2.7$      & 
 $\bm{1.3\pm1.7}$      &    $3.8\pm4.6$      &     $\bm{2.2\pm1.2}$      &    $6.3\pm4.8$      &     $\bm{1.6\pm1.0}$      &    $5.5\pm3.0$      \\
 
 $\mathrm{ATE}_\mathrm{rot} \mathrm{(\degree)}$ &     $\bm{19.7\pm7.8}$      &     $66.2\pm59.5$     &  $\bm{22.8\pm17.2}$     &     $61.1\pm68.1$     &     $\bm{25.3\pm18.4}$     &     $66.9\pm48.9$     &      $\bm{19.4\pm9.5}$     &     $55.8\pm22.4$     \\
 
 $\mathrm{RPE}_\mathrm{trans} \mathrm{(mm)}$ &    $\bm{1.3\pm0.4}$       &    $2.5\pm1.4$      &  $\bm{1.4\pm0.7}$       &    $2.7\pm2.1$      &      $\bm{1.9\pm0.6}$       &    $4.8\pm3.5$      &     $\bm{1.2\pm0.5}$       &    $3.6\pm1.6$      \\
 
 $\mathrm{RPE}_\mathrm{rot} \mathrm{(\degree)}$ &    $\bm{5.9\pm1.7}$       &     $6.4\pm3.5$         &  
 $4.3\pm2.0$            &     $\bm{3.8\pm2.6}$    &   
 $\bm{7.4\pm2.6}$       &     $7.7\pm3.9$         &      $\bm{4.5\pm1.1}$       &     $8.5\pm2.9$         \\
 
 $\mathrm{ARD}_\mathrm{traj}$ &    
 $\bm{0.39\pm 0.17}$     &     $1.73\pm 1.02$     &          $\bm{0.34\pm 0.10}$     &     $2.00\pm 1.82$     &      $\bm{0.38\pm0.14}$      &     $1.58\pm 1.42$     &    $\bm{0.29\pm 0.09}$     &     $1.56\pm 1.20$     \\
 
  $\mathrm{ARD}_\mathrm{frame}$ &    
 $\bm{0.17\pm 0.04}$     &     $1.73\pm 1.02$     &          $\bm{0.17\pm 0.04}$     &     $2.00\pm 1.82$     &      $\bm{0.18\pm0.03}$      &     $1.58\pm 1.42$     &    $\bm{0.15\pm 0.02}$     &     $1.56\pm 1.20$     \\
 
\makecell[l]{$\mathrm{Threshold}_\mathrm{traj}$ \\ $(\theta=1.25)$} &   $\bm{0.39\pm 0.19}$   &     $0.15\pm0.13$     &          $\bm{0.46\pm0.14}$    &     $0.24\pm0.21$     &          $\bm{0.38\pm0.15}$    &     $0.14\pm0.14$     &    $\bm{0.49\pm0.13}$    &     $0.14\pm0.15$     \\
 
\makecell[l]{$\mathrm{Threshold}_\mathrm{frame}$ \\ $(\theta=1.25)$} &   $\bm{0.39\pm 0.19}$   &     $0.15\pm0.13$     &          $\bm{0.46\pm0.14}$    &     $0.24\pm0.21$     &          $\bm{0.38\pm0.15}$    &     $0.14\pm0.14$     &    $\bm{0.49\pm0.13}$    &     $0.14\pm0.15$     \\
  
\makecell[l]{$\mathrm{Threshold}_\mathrm{traj}$ \\ $(\theta=1.25^2)$} &   $\bm{0.70\pm 0.22}$       &    $0.28\pm0.22$      &           $\bm{0.81\pm 0.13}$       &    $0.38\pm0.29$      &         $\bm{0.66\pm0.16}$        &    $0.27\pm0.23$      &    $\bm{0.84\pm0.10}$        &    $0.27\pm0.22$      \\
 
\makecell[l]{$\mathrm{Threshold}_\mathrm{frame}$ \\ $(\theta=1.25^2)$} &   $\bm{0.70\pm 0.22}$       &    $0.28\pm0.22$      &           $\bm{0.81\pm 0.13}$       &    $0.38\pm0.29$      &         $\bm{0.66\pm0.16}$        &    $0.27\pm0.23$      &    $\bm{0.84\pm0.10}$        &    $0.27\pm0.22$      \\
  
 \bottomrule
\end{tabular}
}
\end{table*}

\begin{table*}[!]
\renewcommand{\arraystretch}{1.0}
\caption[Ablation study for the proposed SLAM system on trajectory-related metrics]{\textbf{
Ablation study for the proposed SLAM system on trajectory-related metrics.}\quad
}
\label{tab:dense_slam_ablation_study_trajectory}
\centering
\resizebox{0.65\linewidth}{!}{
\begin{tabular}{cccccllll}
\hline
FMT & FMM & RPT & Local & Global & \multicolumn{1}{c}{$\mathrm{ATE}_\mathrm{trans} \mathrm{(mm)}$} & \multicolumn{1}{c}{$\mathrm{ATE}_\mathrm{rot} \mathrm{(\degree)}$} & \multicolumn{1}{c}{$\mathrm{RPE}_\mathrm{trans} \mathrm{(mm)}$} & \multicolumn{1}{c}{$\mathrm{RPE}_\mathrm{rot} \mathrm{(\degree)}$} \\ \hline
\checkmark & \checkmark & \checkmark & \checkmark & \checkmark & $\bm{1.6\pm1.4}$ & $\bm{22.2\pm15.1}$ & $\bm{1.5\pm0.6}$ & $5.5\pm2.4$ \\
 & \checkmark & \checkmark & \checkmark & \checkmark & $3.4\pm2.7^\text{***}$ & $43.3\pm27.9^\text{***}$ & $2.6\pm1.4^\text{***}$ & $7.3\pm3.0^\text{***}$ \\
 &  & \checkmark & \checkmark & \checkmark & $3.3\pm2.8^\text{***}$ & $40.2\pm23.6^\text{***}$ & $2.6\pm1.2^\text{***}$ & $7.0\pm2.6^\text{***}$ \\
\checkmark & \checkmark &  & \checkmark & \checkmark & $2.7\pm5.5$ & $23.8\pm14.5$ & $2.1\pm3.2$ &  $\bm{5.3\pm2.1}$ \\
\checkmark & \checkmark & \checkmark & \checkmark &  & $2.0\pm1.9^\text{*}$ & $26.8\pm21.2^\text{*}$ & $1.5\pm0.7$ &  $5.5\pm2.4$ \\
\checkmark & \checkmark & \checkmark &  &  & $2.0\pm1.9^\text{*}$ & $25.5\pm18.5^\text{*}$ & $1.5\pm0.7$ & $5.4\pm2.4$ \\ \hline
\end{tabular}
}
\end{table*}

\begin{table*}[!]
\renewcommand{\arraystretch}{1.0}
\caption[Ablation study for the proposed SLAM system on depth-related metrics]{\textbf{
Ablation study for the proposed SLAM system on depth-related metrics.}\quad
}
\label{tab:dense_slam_ablation_study_depth}
\centering
\resizebox{0.85\linewidth}{!}{
\begin{tabular}{cccccllllll}
\hline
FMT & FMM & RPT & Local & Global & \makecell[c]{$\mathrm{ARD}_\mathrm{traj}$} & \makecell[c]{$\mathrm{ARD}_\mathrm{frame}$} &
 \makecell[c]{$\mathrm{Threshold}_\mathrm{traj}$ \\ $(\theta=1.25)$} & \makecell[c]{$\mathrm{Threshold}_\mathrm{frame}$ \\ $(\theta=1.25)$} & 
 \makecell[c]{$\mathrm{Threshold}_\mathrm{traj}$ \\ $(\theta=1.25^2)$} & \makecell[c]{$\mathrm{Threshold}_\mathrm{frame}$ \\ $(\theta=1.25^2)$} \\ \hline
\checkmark & \checkmark & \checkmark & \checkmark & \checkmark &         $0.36\pm0.16$             &             $\bm{0.17\pm0.03}$         &            $0.42\pm0.17$          &            $0.73\pm0.08$          &                  $0.74\pm0.21$    &            $\bm{0.95\pm0.04}$          \\
 & \checkmark & \checkmark & \checkmark & \checkmark &          $0.49\pm0.19^\text{***}$            &          $0.17\pm0.03$            &         $0.29\pm0.16^\text{***}$             &          $0.73\pm0.08$            &          $0.59\pm0.23^\text{***}$            &        $0.95\pm0.04$              \\
 &  & \checkmark & \checkmark & \checkmark &          $0.50\pm0.25^\text{**}$            &           $0.17\pm0.03$           &          $0.32\pm0.17^\text{**}$            &           $\bm{0.74\pm0.08}$           &            $0.61\pm0.24^\text{**}$          &        $0.95\pm0.04$              \\
\checkmark & \checkmark &  & \checkmark & \checkmark &         $\bm{0.35\pm0.15}$             &            $0.17\pm0.03$          &          $\bm{0.43\pm0.17}$            &          $0.73\pm0.08$            &          $\bm{0.76\pm0.18}$            &          $0.95\pm0.04$            \\
\checkmark & \checkmark & \checkmark & \checkmark &  &          $0.36\pm0.16$            &           $0.17\pm0.03$           &          $0.42\pm0.17$            &          $0.73\pm0.08$            &          $0.74\pm0.21$            &          $0.95\pm0.04$            \\
\checkmark & \checkmark & \checkmark &  &  &         $0.35\pm0.16^\text{*}$             &          $0.17\pm0.03$            &           $0.42\pm0.18$           &           $0.73\pm0.08$           &          $0.74\pm0.22$            &           $0.95\pm0.04$           \\ \hline
\end{tabular}
}
\end{table*}

\section{Evaluation Metrics}
The metrics used for camera trajectory evaluation are Absolute Trajectory Error (ATE) and Relative Pose Error (RPE)~\cite{zhang2018trajectory}.
Note that only the frames that are treated as keyframes by the SLAM system will be evaluated in terms of both trajectory error and depth error.
Therefore, synchronization needs to be done to first associate the trajectory estimate with the groundtruth one.
The trajectory estimate will then be spatially aligned with the groundtruth trajectory, where a similarity transform is estimated with the method in~\cite{zhang2018trajectory}.

ATE is used to quantify the whole trajectory and the Root Mean Square Error (RMSE) is used.
The rotation and translation components of this metric are defined as
\begin{equation}
\begin{split}
    &\mathrm{ATE}_\mathrm{rot} = \left(\dfrac{1}{N}\sum_{i=0}^{N-1}{\norm{\log{\left(\bm{R}^\mathrm{ATE}_i\right)}}^2}\right)^{1/2}\quad \text{and} \\
    &\mathrm{ATE}_\mathrm{trans} = \left(\dfrac{1}{N}\sum_{i=0}^{N-1}{\norm{\bm{t}^\mathrm{ATE}_i}^2}\right)^{1/2}\quad \text{,}
\end{split}
\end{equation}
where $\bm{R}^\mathrm{ATE}_i = \Tilde{\bm{R}}_i^\mathrm{wld} \left(\bm{R}_i^\mathrm{wld}\right)^\intercal$ and 
$\bm{t}^\mathrm{ATE}_i = \Tilde{\bm{t}}_i^\mathrm{wld} - \bm{R}_i\bm{t}_i^\mathrm{wld}$.
$\Tilde{\bm{R}}_i^\mathrm{wld}\in\mathrm{SO}\left(3\right)$ and $\Tilde{\bm{t}}_i^\mathrm{wld}\in\mathbb{R}^3$ are the groundtruth rotation and translation components of the $i$\textsuperscript{th} pose in the trajectory, respectively,
while $\bm{R}_i^\mathrm{wld}\in\mathrm{SO}\left(3\right)$ and $\bm{t}_i^\mathrm{wld}\in\mathbb{R}^3$ are the estimated ones.
$N\in\mathbb{R}$ is the number of poses in the synchronized and aligned trajectory estimate.

RPE measures the local accuracy of the trajectory over a fixed frame interval $\Delta\in\mathbb{R}$.
This measures the local drift of the trajectory, which is less affected by the loop closure and emphasizes more on the other components of the system.
The rotation and translation components of this metric are defined as
\begin{equation}\label{eq:dense_slam_rpe_metric}
\begin{split}
    &\mathrm{RPE}_\mathrm{rot} = \left(\dfrac{1}{N - \Delta}\sum_{i=0}^{N-\Delta-1}{\norm{\log{\left(\bm{R}^\mathrm{RPE}_i\right)}}^2}\right)^{1/2}\quad \text{and} \\
    &\mathrm{RPE}_\mathrm{trans} = \left(\dfrac{1}{N-\Delta}\sum_{i=0}^{N-\Delta-1}{\norm{\bm{t}^\mathrm{RPE}_i}^2}\right)^{1/2}\quad \text{,}
\end{split}
\end{equation}
$\bm{R}^\mathrm{RPE}_i\in\mathrm{SO}\left(3\right)$ and $\bm{t}^\mathrm{RPE}_i\in\mathbb{R}^3$ are the rotation and translation components of $\bm{T}^\mathrm{RPE}_i\in\mathrm{SE}\left(3\right)$, respectively;
$\bm{T}^\mathrm{RPE}_i$ is the $i$\textsuperscript{th} RPE matrix, which is defined as
\begin{equation}
    \bm{T}^\mathrm{RPE}_i = \left((\Tilde{\bm{T}}_i^\mathrm{wld})^{-1} \Tilde{\bm{T}}_{i+\Delta}^\mathrm{wld}\right)^{-1} \left((\bm{T}_i^\mathrm{wld})^{-1} \bm{T}_{i+\Delta}^\mathrm{wld}\right)\quad \text{.}
\end{equation}
$\Delta$ in Eq.~\ref{eq:dense_slam_rpe_metric} is set to $7$ for our results;
for ORB-SLAM3, $\Delta$ is set so that the number of original video frames between $\bm{T}_i^\mathrm{wld}$ and $\bm{T}_{i+\Delta}^\mathrm{wld}$ is roughly the same as ours.

To evaluate depth estimates, Absolute Relative Difference (ARD) and Threshold~\cite{yin2018geonet} are used.
Before computing metrics, different pre-processing is applied for two sets of metrics, which are $\mathrm{ARD}_\mathrm{traj}$ and $\mathrm{Threshold}_\mathrm{traj}$, and $\mathrm{ARD}_\mathrm{frame}$ and $\mathrm{Threshold}_\mathrm{frame}$.
For the former, the estimated depth per keyframe is scaled with the scale component in the similarity transform obtained from the trajectory alignment above.
For the latter, each depth estimate is scaled with the median value of ratios between the corresponding groundtruth one and the estimate.
In terms of the definitions of these metrics, ARD is 
\begin{equation}
\mathrm{ARD} = \dfrac{1}{N}\sum_{i = 0}^{N - 1}\dfrac{1}{\vert\Omega_i\vert}\sum_{\bm{x}\in\Omega}{\dfrac{\vert \bm{D}_i\left(\bm{x}\right) - \Tilde{\bm{D}}_i\left(\bm{x}\right)\vert}{\Tilde{\bm{D}}_i\left(\bm{x}\right)}}\quad \text{;}
\end{equation}
Threshold is
\begin{equation}
\begin{split}
\mathrm{Threshold} = \dfrac{1}{N}
&\sum_{i = 0}^{N - 1}\dfrac{1}{\vert\Omega_i\vert}\sum_{\bm{x}\in\Omega} \\
&{\mathbbm{1}\left[\mathrm{max}\left(\dfrac{\bm{D}_i\left(\bm{x}\right)}{\Tilde{\bm{D}}_i\left(\bm{x}\right)}, \dfrac{\Tilde{\bm{D}}_i\left(\bm{x}\right)}{\bm{D}_i\left(\bm{x}\right)}\right) < \theta\right]}\quad \text{.}
\end{split}
\end{equation}
Note that $\Omega_i$ here is the region where both scaled depth estimate $\bm{D}_i\in\mathbb{R}^{1\times H\times W}$ and groundtruth depth $\Tilde{\bm{D}}_i\in\mathbb{R}^{1\times H\times W}$, for the $i$\textsuperscript{th} synchronized keyframe, have valid depths;
$\theta\in\mathbb{R}$ is the threshold used to determine if the depth ratio between the estimate and groundtruth is small enough.

\section{Cross-Subject Evaluation Additional Results}
To evaluate the performance of the SLAM system on endoscopic videos from unseen subjects, we run a cross-validation study.
Four models are trained with different train/test splits on the $11$ subjects in total.
With subjects named as consecutive numbers, the test splits for $4$ models are $\{1, 2, 3\}$, $\{4, 5, 6\}$, $\{7, 8, 11\}$, and $\{8, 9, 10\}$, and the train splits for each model are the subjects left.
The evaluation metrics, as reported in Table~\ref{tab:dense_slam_cross_evaluation}, are averaged over all the sequences within the corresponding test split for each trained model.
Besides, we also compare against a state-of-the-art feature-based SLAM system, ORB-SLAM3~\cite{ORBSLAM32020}, which we evaluate on all videos at once and use the same set of metrics for evaluation.

\section{Ablation Study Results}
Table~\ref{tab:dense_slam_ablation_study_trajectory} and \ref{tab:dense_slam_ablation_study_depth} show the ablation study results on trajectory-related and depth-related metrics, respectively.
As can be seen, FM has a large impact on both trajectory and trajectory-scaled depth metrics;
RP mainly affects trajectory metrics;
the \emph{Loop Closure} module mainly affects the trajectory metrics $\mathrm{ATE}_\mathrm{trans}$ and $\mathrm{ATE}_\mathrm{rot}$.

\bibliographystyle{IEEEtran}
\bibliography{Bibliography}